\documentclass[11pt]{article}
\usepackage[a4paper]{geometry}
\usepackage{clic2016}
\usepackage{times}
\usepackage{url}
\usepackage{latexsym}
\usepackage{graphicx}
\usepackage{booktabs}
\usepackage{xcolor}
\usepackage{enumitem}

\title{When silver glitters more than gold:\\Bootstrapping an Italian part-of-speech tagger for Twitter}

\author{Barbara Plank\\
University of Groningen\\
The Netherlands\\
  {\tt b.plank@rug.nl} \\\And
  Malvina Nissim\\ 
University of Groningen\\
The Netherlands\\
  {\tt m.nissim@rug.nl}\\}

\date{}

\begin{document}
\maketitle
\begin{abstract}
  \textbf{English.} 
  We bootstrap a state-of-the-art part-of-speech tagger to tag Italian Twitter data, in the context of the Evalita~2016 PoSTWITA shared task. We show that training the tagger on native Twitter data enriched with little amounts of specifically selected gold data and additional silver-labelled data scraped from Facebook, yields better results than using large amounts of manually annotated data from a mix of genres.
\end{abstract}

\begin{abstract-alt}
 \textrm{\bf{Italiano.}} Nell'ambito della campagna di valutazione PoSTWITA di Evalita~2016, addestriamo due modelli che differiscono nel grado di supervisione  in fase di training. Il modello addestrato con due cicli di bootstrapping usando post da Facebook, e che quindi impara anche da etichette ``silver", ha una performance superiore alla versione supervisionata che usa solo dati annotati manualmente. Discutiamo l'importanza della scelta dei dati di training e development.
 \end{abstract-alt}

\section{Introduction}

The emergence and abundance of social media texts has prompted the urge to develop tools that are able to process  language which is often non-conventional, both in terms of lexicon as well as grammar. Indeed, models trained on standard newswire data heavily suffer when used on data from a different language variety, especially Twitter~\cite{McClosky:ea:10,Foster:ea:11,Gimpel:ea:2011,Plank:2016:KONVENS}. 

As a way to equip microblog processing with efficient tools, two ways of developing Twitter-compliant models have been explored. One option is to transform Twitter language back to what pre-trained models already know via normalisation operations, so that existing tools are more successful on such different data. The other option is to create \textit{native} models by training them on  labelled Twitter data. The drawback of the first option is that it's not clear what norm to target: ``what is standard language?"~\cite{eisenstein:2013:bad,Plank:2016:KONVENS}, and implementing normalisation procedures requires quite a lot of manual intervention and subjective decisions. The drawback of the second option is that manually annotated Twitter data isn't readily available, and it is costly to produce. 

In this paper, we report on our participation in PoSTWITA\footnote{{http://corpora.ficlit.unibo.it/PoSTWITA/}}, the EVALITA~2016 shared task on Italian Part-of-Speech (POS) tagging for Twitter~\cite{postwita2016overview}. We emphasise an approach geared to building a \textit{single model} (rather than an ensemble) based on weakly supervised learning, thus favouring (over normalisation) the aforementioned second option of learning \textit{invariant representations}, also for theoretical reasons. We address the bottleneck of acquiring manually annotated data  by suggesting and showing that a semi-supervised approach that mainly focuses on tweaking data selection within a bootstrapping setting can be successfully pursued for this task. Contextually, we show that large amounts of manually annotated data might not be helpful if data isn't ``of the right kind''.

\section{Data selection and bootstrapping}\label{sec:approach}

In adapting a POS tagger to Twitter, we mainly focus on ways of selectively enriching the training set with additional data. Rather than simply adding large amounts of existing annotated data, we investigate ways of selecting smaller amounts of more appropriate training instances, possibly even tagged with silver rather than gold labels.  As for the model itself, we simply take an off-the-shelf tagger, namely a bi-directional Long Short-Term Memory (bi-LSTM) model \cite{plank:ea:2016}, which we use with default parameters (see Section~\ref{sec:bilstm}) apart from initializing it with Twitter-trained embeddings (Section~\ref{sec:setup}). 

Our first model is trained on the PoSTWITA training set plus additional gold data selected according to two criteria (see below: \textit{Two shades of gold}). This model is used to tag a collection of Facebook posts in a bootstrapping setting with two cycles (see below: \textit{Bootstrapping via Facebook}). 
The rationale behind using Facebook as \textit{not-so-distant} source when targeting Twitter is the following: many Facebook posts of public, non-personal pages resemble tweets in style, because of brevity and the use of  hashtags. However, differently from random tweets, they are usually correctly formed grammatically and spelling-wise, and often provide more context, which allows for more accurate tagging.

\paragraph{Two shades of gold}
We used the Italian portion of the latest release (v1.3) of the Universal Dependency (UD) dataset~\cite{ud1.3}, from which we extracted two subsets, according to two different criteria. First, we selected data on the basis of its \textit{origin}, trying to match the Twitter training data as close as possible. For this reason, we used the Facebook subportion (\texttt{UD\_FB}). These are 45 sentences that presumably stem from the Italian Facebook help pages and contain questions and short answers.\footnote{These are labelled as \texttt{4-FB} in the comment section of UD. Examples include: \textit{Prima di effettuare la registrazione. \`{E} vero che Facebook sar\`{a} a pagamento?}} Second, by looking at the confusion matrix of one of the initial models, we saw that the model's performance was especially poor for cliticised verbs and interjections, tags that are also infrequent in the training set (Table~\ref{tab:postwita}). Therefore, from the Italian UD portion we selected \textit{any} data (in terms of origin/genre) which contained the \texttt{VERB\_CLIT} or \texttt{INTJ} tag, with the aim to boost the identification of these categories. We refer to this set of 933 sentences as \texttt{UD\_verb\_clit+intj}.

\parskip 0.2cm

\begin{table}
\caption{Statistics on the additional datasets.\label{tbl:stats}
}
\resizebox{\columnwidth}{!}{
\begin{tabular}{llrr}
\toprule
\textbf{Data}     & \textbf{Type} & \textbf{Sents} & \textbf{Tokens}\\
\midrule
\texttt{UD\_FB} & gold & 45 & 580\\

\texttt{UD\_verb\_clit+intj} & gold & 933 & 26k\\
\texttt{FB} (all, iter 1)   & silver & 2243 & 37k\\
\texttt{FB} (all, iter 2)   & silver & 3071 & 47k\\
\midrule
\textbf{Total added data}   & gold+silver & 4049 & 74k\\
\bottomrule
\end{tabular}
} 
\end{table}

\paragraph{Bootstrapping via Facebook}

We augmented our training set with silver-labelled data. With our best model trained on the original task data plus \texttt{UD\_verb\_clit+intj} and \texttt{UD\_FB}, we tagged a collection of Facebook posts, added those to the training pool, and retrained our tagger. We used two iterations of indelible self-training~\cite{abney:2007}, i.e., adding automatically tagged data where labels do not change once added. Using the Facebook API through the Facebook-sdk python library\footnote{{https://pypi.python.org/pypi/facebook-sdk}}, we scraped an average of 100 posts for each of the following pages, selected on the basis of our intuition and on reasonable site popularity:

\begin{small}
\begin{itemize}[noitemsep]
\item sport: {\small \texttt{corrieredellosport}}
\item news: {\small \texttt{Ansa.it}, \texttt{ilsole24ore}, \texttt{lastampa.it}}
\item politics: {\small \texttt{matteorenziufficiale}}
\item entertainment: {\small \texttt{novella2000}, \texttt{alFemminile}}
\item travel: {\small \texttt{viaggiart}}
\end{itemize}
\end{small}

\noindent We included a second cycle of bootstrapping,  scraping a few more Facebook pages ({\small \texttt{soloGossip.it}, \texttt{paesionline}, \texttt{espressonline}, \texttt{LaGazzettaDelloSport}}, again with an average of 100 posts each), and tagging the posts with the model that had been re-trained on the original training set plus the first round of Facebook data with silver labels (we refer to the whole of the automatically-labelled Facebook data as \texttt{FB\_silver}). \texttt{FB\_silver} was added to the training pool to train the final model. Statistics on the obtained data are given in Table~\ref{tbl:stats}.\footnote{Due to time constraints we did not add further iterations; we cannot judge if we already reached a performance plateau.}

\section{Experiments and Results}

In this section we describe how we developed the two models of the final submission, including all preprocessing decisions. We highlight the importance of choosing an adequate development set to identify promising directions.

\subsection{Experimental Setup}\label{sec:setup}
\paragraph{PoSTWITA data}
In the context of PoSTWITA, training data was provided to all participants in the form of manually labelled tweets. The tags comply with the UD tagset, with a couple of modifications due to the specific genre (emoticons are labelled with a dedicated tag, for example), and subjective choices in the treatment of some morphological traits typical of Italian. Specifically, cli\-tics and articulated prepositions are treated as one single form (see below: \textit{UD fused forms}). The training set contains 6438 tweets, for a total of ca. 115K tokens. The distribution of tags together with  examples is given in Table~\ref{tab:postwita}.  The test set comprises 301 tweets (ca. 4800 tokens).

\begin{table}
\caption{\label{tab:postwita}Tag distribution in the original trainset.}
\begin{scriptsize}
\begin{tabular}{llrl} \bottomrule
\textbf{Tag} & \textbf{Explanation} & \textbf{\#Tokens} & \textbf{Example} \\
\hline
\texttt{NOUN} & noun & 16378 & cittadini \\
\texttt{PUNCT} & punctuation & 14513 & ? \\
\texttt{VERB} & verb & 12380 & apprezzo \\
\texttt{PROPN} & proper noun & 11092 & Ancona\\
\texttt{DET} & determiner & 8955 & il \\
\texttt{ADP} & preposition & 8145 & per \\
\texttt{ADV} & adverb & 6041 & sempre \\
\texttt{PRON} & pronoun & 5656 & quello \\
\texttt{ADJ} & adjective & 5494 & mondiale \\
\texttt{HASHTAG} & hashtag & 5395 & \#manovra \\
\texttt{ADP\_A} & articulated prep & 4465 & nella \\
\texttt{CONJ} & coordinating conj & 2876 & ma \\
\texttt{MENTION} & mention & 2592 & @InArteMorgan\\
\texttt{AUX} & auxiliary verb & 2273 & potrebbe \\
\texttt{URL} & url & 2141 & {{http://t.co/La3opKcp}}\\
\texttt{SCONJ} & subordinating conj& 1521 & quando \\
\texttt{INTJ} & interjection & 1404 & fanculo \\
\texttt{NUM} & number & 1357 & 23\% \\
\texttt{X} & anything else & 776 & s...	\\
\texttt{EMO} & emoticon & 637 &  \includegraphics[scale=.17]{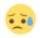} \\
\texttt{VERB\_CLIT} & verb+clitic & 539 & vergognarsi \\
\texttt{SYM} & symbol & 334 & $\rightarrow$\\
{\texttt{PART}} & particle & 3 & 's  \\
\bottomrule
\end{tabular}
\end{scriptsize}
\end{table}

\paragraph{UD fused forms} 
In the UD scheme for Italian, articulated prepositions (\texttt{ADP\_A}) and cliticised verbs (\texttt{VERB\_CLIT}) are annotated as separate word forms, while in PoSTWITA the original word form (e.g., `alla' or `arricchirsi') is annotated as a whole. In order to get the PoSTWITA \texttt{ADP\_A} and \texttt{VERB\_CLIT} tags for these fused word forms from UD, we adjust the UCPH \texttt{ud-conversion-tools}\footnote{{https://github.com/coastalcph/ud-conversion-tools}}~\cite{agic:ea:2016} that propagates head POS information up to the original form.

\paragraph{Pre-processing of unlabelled data} For the Facebook data, we use a simplistic off-the-shelf rule-based tokeniser that segments sentences by punctuation and tokens by whitespace.\footnote{{https://github.com/bplank/multilingualtokenizer}} We normalise URLs to a single token (\texttt{http://www.someurl.org}) and add a rule for smileys. Finally, we remove sentences from the Facebook data were more than 90\% of the tokens are in all caps. Unlabelled data used for embeddings is preprocessed only with normalisation of usernames and URLs.

\paragraph{Word Embeddings}
We induced word embeddings from 5 million Italian tweets (\textsc{Twita}) from Twita~\cite{twita}. Vectors were created using \texttt{word2vec} \cite{mikolov2013distributed} with default parameters, except for the fact that we set the dimensions to 64, to match the vector size of the multilingual (\textsc{Poly}) embeddings \cite{al2013polyglot} used by \newcite{plank:ea:2016}. We dealt with unknown words by adding a ``UNK'' token computing the mean vector of three infrequent words (``vip!",``cuora", ``White").

\begin{figure}[h]\centering
 \includegraphics[scale=.22]{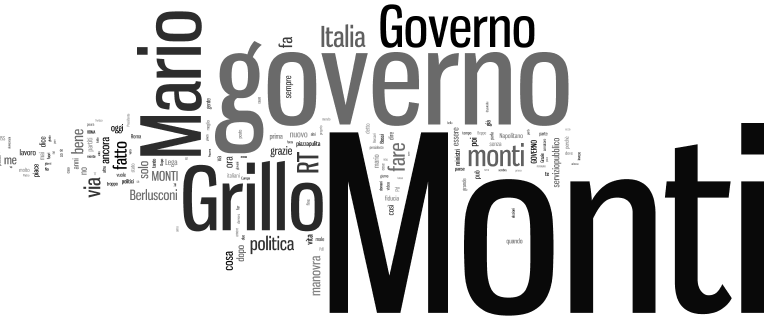}
 \caption{Word cloud from the training data.}
 \label{fig:wordle}
\end{figure}

\paragraph{Creation of a \textit{realistic} internal development set} The original task data is distributed as a single training file. 
In initial experiments we saw that performance varied considerably for different random subsets. This was due to a large bias towards tweets about `Monti' and `Grillo', see Figure~\ref{fig:wordle}, but also because of duplicate tweets. We opted to create \textit{the most difficult} development set possible. This development set was achieved by removing duplicates, and randomly selecting a subset of tweets that do not mention `Grillo' or `Monti' while maximizing out-of-vocabulary (OOV) rate with respect to the training data.  Hence, our internal development set consisted of 700 tweets with an OOV approaching 50\%. This represents a more realistic testing scenario. Indeed, the baseline (the basic bi-LSTM model), dropped  from 94.37 to 92.41 computed on the earlier development set were we had randomly selected 1/5 of the data, with an OOV of 45\% (see Table~\ref{tbl:resultsXlime}).

\subsection{Model}

\label{sec:bilstm} 
\begin{figure}
\centering
\includegraphics[scale=.3]{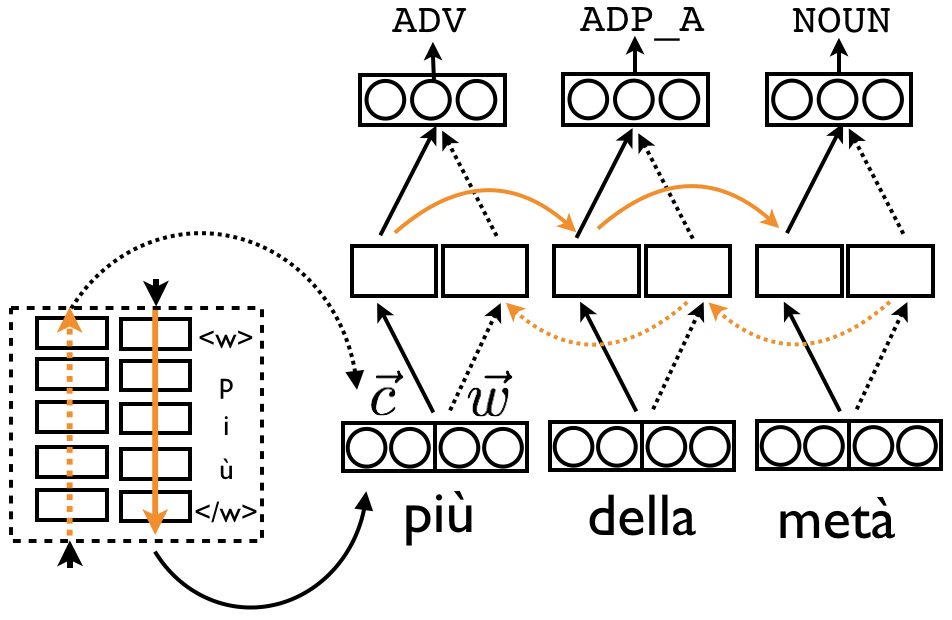}
\caption{Hierarchical bi-LSTM model using word $\vec{w}$ and character $\vec{c}$ representations.}
\label{fig:model}
\end{figure}

The bidirectional Long Short-Term Memory model \texttt{bilty}\footnote{{https://github.com/bplank/bilstm-aux}} is illustrated in Figure~\ref{fig:model}. It is a context bi-LSTM taking as input word embeddings $\vec{w}$. Character embeddings $\vec{c}$ are incorporated via a
hierarchical bi-LSTM using a sequence bi-LSTM at the lower level~\cite{ballesteros:ea:2015,plank:ea:2016}. The character representation is concatenated with the (learned) word
embeddings $\vec{w}$ to form the input to the context bi-LSTM at the upper layers. 
We took default parameters, i.e., character embeddings set to 100, word embeddings set to 64, 20 iterations of training using Stochastic Gradient Descent,
a single bi-LSTM layer and regularization using Gaussian noise with $\sigma=0.2$ (\texttt{cdim 100, trainer sgd, indim 64, iters 20, h\_layer 1, sigma 0.2}). 
The model has been shown to achieve state-of-the-art performance on a range of languages, where the incorporation of character information was particularly effective~\cite{plank:ea:2016}. With these features and settings we train two models on different training sets.

\paragraph{\textsc{GoldPick}} \texttt{bilty} with pre-initialised \textsc{Twita} embeddings, trained on the PoSTWITA training set plus selected gold data (\texttt{UD\_FB} + \texttt{UD\_verb\_clit+intj}).

\paragraph{\textsc{SilverBoot}}  a bootstrapped version of \textsc{GoldPick}, where {\texttt{FB\_silver}} (see~Section~\ref{sec:approach}) is also added to the training pool, which thus includes both gold and silver data.

\subsection{Results on test data}
Participants were allowed to submit one official, and one additional (unofficial) run. Because on development data \textsc{SilverBoot} performed better than \textsc{GoldPick}, we selected the former for our official submission and the latter for the unofficial one, making it thus also possible to assess the specific contribution of bootstrapping to performance.

Table~\ref{tbl:resultsTest} shows the results on the official test data for both our models and \textsc{TnT}~\cite{Brants:00}.  
The results show that adding bootstrapped silver data outperforms the model trained on gold data alone. 
The additional training data included in \textsc{SilverBoot} reduced the OOV rate for the testset to 41.2\% (compared to 46.9\% with respect to the original PoSTWITA training set).
Note that, on the original randomly selected development set the results were less indicative of the contribution of the silver data (see Table~\ref{tbl:resultsXlime}), showing the importance of a carefully selected development set. 

\begin{table}\centering
\caption{\label{tbl:resultsTest}
Results on the official test set. \textsc{Best} is the highest performing system at PoSTWITA.}
\resizebox{0.9\columnwidth}{!}{
\begin{tabular}{lr}
   \toprule
\textbf{System}    & \textbf{Accuracy} \\
\midrule
\textsc{Best} & 93.19 \\
\textsc{SilverBoot} (official)& 92.25\\
\textsc{GoldPick} (unofficial) & 91.85 \\ 
\textsc{TnT} (on \textsc{PoSTWITA} train) & 84.83\\
\textsc{TnT} (on \textsc{SilverBoot} data) & 85.52\\
\bottomrule
\end{tabular}
}
\end{table}

\section{What didn't work}

In addition to what we found to boost the tagger's performance, we also observed what didn't yield any improvements, and in some case even lowered global accuracy. What we experimented with was triggered by intuition and previous work, as well as what we had already found to be successful, such as selecting additional data to make up for under-represented tags in the training set. However, everything we report in this section turned out to be either pointless or detrimental.

\paragraph{More data} 
We added to the training data \textit{all} (train, development, and
test) sections from the Italian part of UD1.3. While training on
selected gold data (978 sentences) yielded 95.06\% accuracy, adding
all of the UD-data (12k sentences of newswire, legal and wiki texts) yielded a
disappointing 94.88\% in initial experiments (see Table~\ref{tbl:resultsXlime}), also considerably
slowing down training.

Next, we tried to add more Twitter data from \textsc{xLIME}, a
publicly available corpus with multiple layers of manually assigned
labels, including POS tags, for a total of ca. 8600 tweets and 160K
tokens~\cite{rei:ea:2016}.
The data isn't provided as a single gold standard file but in the form
of separate annotations produced by different judges, so that we used
MACE~\cite{Hovy:ea:13} to adjudicate divergences. Additionally, the
tagset is slightly different from the UD set, so that we had to
implement a mapping. The results in Table~\ref{tbl:resultsXlime} show
that adding all of the \textsc{xLIME} data declines performance,
despite careful preprocessing to map the tags and resolve annotation
divergences.

\begin{table}
\caption{Results on internal development set.\label{tbl:resultsXlime}
}
\centering
\begin{tabular}{lr}
   \toprule
   \textbf{System}    & \textbf{Accuracy} \\
\midrule
\multicolumn{2}{c}{Internal dev (prior) OOV: 45\%}\\
\hline
\textsc{Baseline} (w/o emb) & 94.37\\
+\textsc{Poly} emb & 94.15\\
+\textsc{Twita} emb & \textbf{94.69}\\ 
\midrule
\multicolumn{2}{c}{\textsc{Baseline}+\textsc{Twita} emb}\\
+{Morphit!} coarse MTL & 94.61\\ 
+{Morphit!} fine MTL & 94.68\\
\midrule
+{UD all} & 94.88\\
+{gold}-picked & 95.06\\
+{gold}-picked+silver (1st round) & \textbf{95.08}\\
\bottomrule
\multicolumn{2}{c}{Internal dev (realistic) OOV: 50\%}\\
\midrule
\textsc{Baseline} (incl. \textsc{Twita} emb) & 92.41\\
+{gold}  (\textsc{GoldPick}) & 93.19\\
+{gold}+{silver} (\textsc{SilverBoot}) & \textbf{93.42}\\
\multicolumn{2}{c}{adding more gold (Twitter) data:}\\
+\textsc{Xlime adjudicated}  (48) & 92.58\\
+\textsc{Xlime single annot.} & 91.67\\
+\textsc{Xlime all} (8k) & 92.04\\ 
\bottomrule
\end{tabular}
\end{table}

\paragraph{More tag-specific data}
From the matrix computed on the dev set, it emerged that the most confused categories were \texttt{NOUN} and \texttt{PROPN}. Following the same principle that led us to add \texttt{UD\_verb\_clit+intj}, we tried to reduce such confusion by providing additional training data containing proper nouns. 
This did not yield any improvements, neither in terms of global accuracy, nor in terms of precision and recall of the two tags.

\paragraph{Multi-task learning}
Multi-task learning (MTL) \cite{caruana:97}, namely a learning setting where more than one task is learnt at the same time, has been shown to improve performance for several NLP tasks \cite{collobert2011natural,bordes2012joint,liu2015representation}. Often, what is learnt is one main task and, additionally, a number of auxiliary tasks, where the latter should help the model converge better and overfit less on the former. In this context, the additional signal we use to support the learning of each token's POS tag is the token's degree of ambiguity.  Using the information stored in \textit{Morph-it!}, a lexicon of Italian inflected forms with their lemma and morphological features \cite{morphit}, we obtained the \textit{number} of all different tags potentially associated to each token. Because the \textit{Morph-it!} labels are highly fine-grained
we derived two different ambiguity scores, one on the original and one on coarser tags. 
In neither case the additional signal contributed to the tagger's performance, but we have not explored this direction fully and leave it for future investigations.

\section{Conclusions}

The main conclusion we draw from the experiments in this paper is that \textit{data selection matters}, not only for training but also while developing for taking informed decisions. 
Indeed, only after creating a carefully designed internal development set we obtained stronger evidence of the contribution of silver data which is also reflected in the official results.
We also observe that choosing less but more targeted data is more effective. For instance, 
 \textsc{Twita} embeddings contribute more than generic \textsc{Poly} embeddings which were trained on substantially larger amounts of Wikipedia data. 
Also,  just blindly adding training data does not help. We have seen that using the whole of the UD corpus is not beneficial to performance when compared to a small amount of selected gold data, both in terms of origin and labels covered. 
 Finally, and most importantly, we have found that adding little amounts of not-so-distant silver data obtained via bootstrapping resulted in our best model. 
  
We believe the low performance observed when adding xLIME data is likely due to the non-correspondence of tags in the two datasets, which required a heuristic-based mapping. While this is only a speculation that requires further investigation, it seems to indicate that exploring semi-supervised strategies is preferrable to producing idiosyncratic or project-specific gold annotations.

\paragraph*{Acknowledgments}
We thank the CIT of the University of Groningen for providing access to the Peregrine HPC cluster. 
Barbara Plank acknowledges NVIDIA corporation for support.


\begin{thebibliography}{}

\bibitem[\protect\citename{Abney}2007]{abney:2007}
Steven Abney.
\newblock 2007.
\newblock {\em Semisupervised learning for computational linguistics}.
\newblock CRC Press.

\bibitem[\protect\citename{Agi{\'c} \bgroup et al.\egroup }2016]{agic:ea:2016}
{\v Z}eljko Agi{\'c}, Anders Johannsen, Barbara Plank, H{\'e}ctor
  Mart{\'i}nez~Alonso, Natalie Schluter, and Anders S\o{}gaard.
\newblock 2016.
\newblock Multilingual projection for parsing truly low-resource languages.
\newblock {\em Transactions of the Association for Computational Linguistics
  (TACL)}, 4:301--312.

\bibitem[\protect\citename{Al-Rfou \bgroup et al.\egroup }2013]{al2013polyglot}
Rami Al-Rfou, Bryan Perozzi, and Steven Skiena.
\newblock 2013.
\newblock Polyglot: Distributed word representations for multilingual {NLP}.
\newblock {\em arXiv preprint arXiv:1307.1662}.

\bibitem[\protect\citename{Ballesteros \bgroup et al.\egroup
  }2015]{ballesteros:ea:2015}
Miguel Ballesteros, Chris Dyer, and Noah~A. Smith.
\newblock 2015.
\newblock Improved transition-based parsing by modeling characters instead of
  words with lstms.
\newblock In {\em EMNLP}.

\bibitem[\protect\citename{Basile and Nissim}2013]{twita}
Valerio Basile and Malvina Nissim.
\newblock 2013.
\newblock Sentiment analysis on italian tweets.
\newblock In {\em Proceedings of the 4th Workshop on Computational Approaches
  to Subjectivity, Sentiment and Social Media Analysis}, pages 100--107.

\bibitem[\protect\citename{Bordes \bgroup et al.\egroup }2012]{bordes2012joint}
Antoine Bordes, Xavier Glorot, Jason Weston, and Yoshua Bengio.
\newblock 2012.
\newblock Joint learning of words and meaning representations for open-text
  semantic parsing.
\newblock In {\em AISTATS}, volume 351, pages 423--424.

\bibitem[\protect\citename{Brants}2000]{Brants:00}
Thorsten Brants.
\newblock 2000.
\newblock Tnt: a statistical part-of-speech tagger.
\newblock In {\em ANLP}.

\bibitem[\protect\citename{Caruana}1997]{caruana:97}
Rich Caruana.
\newblock 1997.
\newblock Multitask learning.
\newblock {\em Machine Learning}, 28:41--75.

\bibitem[\protect\citename{Collobert \bgroup et al.\egroup
  }2011]{collobert2011natural}
Ronan Collobert, Jason Weston, L{\'e}on Bottou, Michael Karlen, Koray
  Kavukcuoglu, and Pavel Kuksa.
\newblock 2011.
\newblock Natural language processing (almost) from scratch.
\newblock {\em Journal of Machine Learning Research}, 12(Aug):2493--2537.

\bibitem[\protect\citename{Eisenstein}2013]{eisenstein:2013:bad}
Jacob Eisenstein.
\newblock 2013.
\newblock What to do about bad language on the internet.
\newblock In {\em Proceedings of the Annual Conference of the North American
  Chapter of the Association for Computational Linguistics (NAACL)}, pages
  359--369, Atlanta.

\bibitem[\protect\citename{Foster \bgroup et al.\egroup }2011]{Foster:ea:11}
Jennifer Foster, Ozlem Cetinoglu, Joachim Wagner, Josef~Le Roux, Joakim Nivre,
  Deirde Hogan, and Josef van Genabith.
\newblock 2011.
\newblock {From news to comments: Resources and benchmarks for parsing the
  language of Web 2.0}.
\newblock In {\em IJCNLP}.

\bibitem[\protect\citename{Gimpel \bgroup et al.\egroup }2011]{Gimpel:ea:2011}
Kevin Gimpel, Nathan Schneider, Brendan O'Connor, Dipanjan Das, Daniel Mills,
  Jacob Eisenstein, Michael Heilman, Dani Yogatama, Jeffrey Flanigan, and
  Noah~A. Smith.
\newblock 2011.
\newblock {Part-of-Speech Tagging for Twitter: Annotation, Features, and
  Experiments}.
\newblock In {\em Proceedings of ACL}.

\bibitem[\protect\citename{Hovy \bgroup et al.\egroup }2013]{Hovy:ea:13}
Dirk Hovy, Taylor Berg-Kirkpatrick, Ashish Vaswani, and Eduard Hovy.
\newblock 2013.
\newblock Learning whom to trust with {MACE}.
\newblock In {\em NAACL}.

\bibitem[\protect\citename{Liu \bgroup et al.\egroup
  }2015]{liu2015representation}
Xiaodong Liu, Jianfeng Gao, Xiaodong He, Li~Deng, Kevin Duh, and Ye-Yi Wang.
\newblock 2015.
\newblock Representation learning using multi-task deep neural networks for
  semantic classification and information retrieval.
\newblock In {\em Proc. NAACL}.

\bibitem[\protect\citename{McClosky \bgroup et al.\egroup
  }2010]{McClosky:ea:10}
David McClosky, Eugene Charniak, and Mark Johnson.
\newblock 2010.
\newblock {Automatic domain adaptation for parsing}.
\newblock In {\em NAACL-HLT}.

\bibitem[\protect\citename{Mikolov and Dean}2013]{mikolov2013distributed}
T~Mikolov and J~Dean.
\newblock 2013.
\newblock Distributed representations of words and phrases and their
  compositionality.
\newblock {\em Advances in neural information processing systems}.

\bibitem[\protect\citename{Nivre \bgroup et al.\egroup }2016]{ud1.3}
Joakim Nivre et al.
\newblock 2016.
\newblock Universal dependencies 1.3.
\newblock {LINDAT}/{CLARIN} digital library at the Institute of Formal and
  Applied Linguistics, Charles University in Prague.

\bibitem[\protect\citename{Plank \bgroup et al.\egroup }2016]{plank:ea:2016}
Barbara Plank, Anders S{\o}gaard, and Yoav Goldberg.
\newblock 2016.
\newblock Multilingual part-of-speech tagging with bidirectional long
  short-term memory models and auxiliary loss.
\newblock In {\em ACL}.

\bibitem[\protect\citename{Plank}2016]{Plank:2016:KONVENS}
Barbara Plank.
\newblock 2016.
\newblock What to do about non-standard (or non-canonical) language in {NLP}.
\newblock In {\em KONVENS}.

\bibitem[\protect\citename{Rei \bgroup et al.\egroup }2016]{rei:ea:2016}
Luis Rei, Dunja Mladenic, and Simon Krek.
\newblock 2016.
\newblock A multilingual social media linguistic corpus.
\newblock In {\em Conference of CMC and Social Media Corpora for the
  Humanities}.

\bibitem[\protect\citename{Tamburini \bgroup et al.\egroup
  }2016]{postwita2016overview}
Fabio Tamburini, Cristina Bosco, Alessandro Mazzei, and Andrea Bolioli.
\newblock 2016.
\newblock {Overview of the EVALITA 2016 Part Of Speech on TWitter for ITAlian
  Task}.
\newblock In Pierpaolo Basile, Franco Cutugno, Malvina Nissim, Viviana Patti,
  and Rachele Sprugnoli, editors, {\em {Proceedings of the 5th Evaluation
  Campaign of Natural Language Processing and Speech Tools for Italian (EVALITA
  2016)}}. {aAcademia University Press}.

\bibitem[\protect\citename{Zanchetta and Baroni}2005]{morphit}
Eros Zanchetta and Marco Baroni.
\newblock 2005.
\newblock {Morph-it! A free corpus-based morphological resource for the Italian
  language}.
\newblock {\em Corpus Linguistics 2005}, 1(1).

\end{thebibliography}

\end{document}